\documentclass[10pt,twocolumn,letterpaper]{article}

\usepackage[]{cvpr}

\usepackage[pagebackref=true,breaklinks=true,colorlinks,bookmarks=false]{hyperref}

\usepackage[utf8]{inputenc} % allow utf-8 input
\usepackage[T1]{fontenc}    % use 8-bit T1 fonts
\usepackage{hyperref}       % hyperlinks
\usepackage{url}            % simple URL typesetting
\usepackage{booktabs}       % professional-quality tables
\usepackage{amsfonts}       % blackboard math symbols
\usepackage{nicefrac}       % compact symbols for 1/2, etc.
\usepackage{microtype}      % microtypography
\usepackage{xcolor}         % colors
\usepackage{algorithm}
\usepackage{algpseudocode}
\usepackage{amsmath}
\usepackage{graphicx}
\title{Zero-shot racially balanced dataset generation using an existing biased StyleGAN2}

\author{Anubhav Jain, Nasir Memon, Julian Togelius\\
New York University\\
{\tt\small \{aj3281, nm1214, jt125\}@nyu.edu}
}

\begin{document}

\maketitle

\begin{abstract}
Facial recognition systems have made significant strides thanks to data-heavy deep learning models, but these models rely on large privacy-sensitive datasets. Further, many of these datasets lack diversity in terms of ethnicity and demographics, which can lead to biased models that can have serious societal and security implications. To address these issues, we propose a methodology that leverages the biased generative model StyleGAN2 to create demographically diverse images of synthetic individuals. The synthetic dataset is created using a novel evolutionary search algorithm that targets specific demographic groups. By training face recognition models with the resulting balanced dataset containing 50,000 identities per race (13.5 million images in total), we can improve their performance and minimize biases that might have been present in a model trained on a real dataset.
\end{abstract}

\section{Introduction}
Facial recognition systems that were once based on handcrafted features have now achieved human-level performance with the assistance of deep learning models. However, this transition has resulted in the accumulation of large privacy-sensitive datasets that are costly to collect and pose several issues. One major issue is that these large datasets often lack ethnic and demographic diversity, which causes deep facial recognition models to suffer from similar biases. Ensuring the collection of highly diverse image datasets is not only difficult but also expensive. Another issue is that many countries have recognized biometric data privacy as a fundamental right and have regulated its collection and usage by law \cite{voigt2017eu,india_supreme,de2018guide}. This makes it challenging to collect data from a large number of users and raises privacy concerns.  Companies like Facebook \cite{facebook}, Google \cite{google}, and Shutterfly \cite{shutterfly} have faced scrutiny for their usage of facial images of users under the BIPA law.

\begin{figure}
    \centering
    \includegraphics[width=\linewidth]{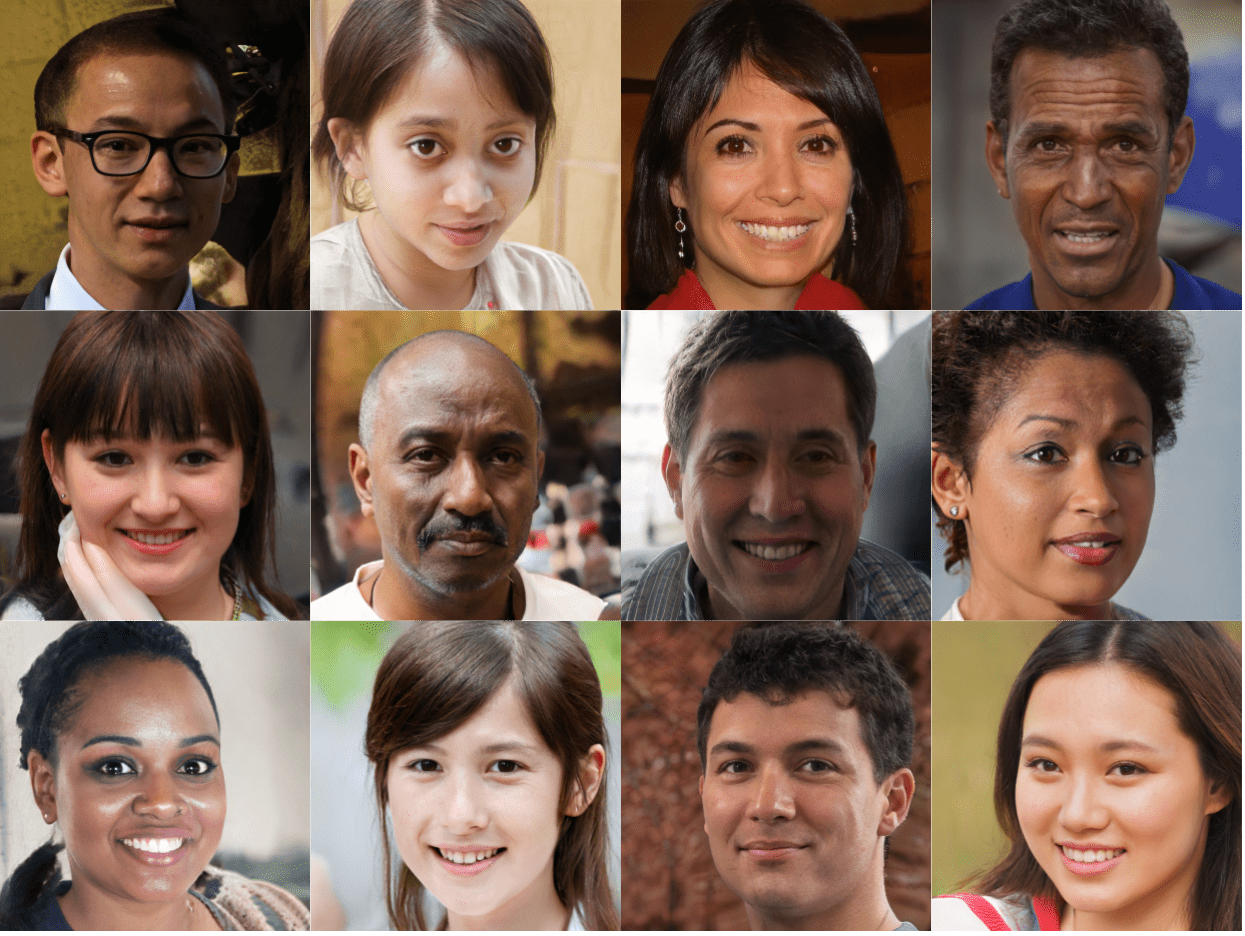}
    \caption{Examples of images generated for different ethnicities using our approach.}
    \label{fig:examples }
\end{figure}

This work presents an approach to address the issues of bias and data privacy in facial recognition models by leveraging advancements in image generation. Generative models offer a cost and time-effective alternative to manual data collection and annotation. They provide better control over environmental conditions, lighting, occlusions, camera angles, and backgrounds, enabling the training of more robust and adaptable models. Additionally, synthetic datasets do not contain any personally identifiable information, thus providing crucial privacy protection.

However, there is a major challenge in using existing generative models as they are highly biased. For instance, when randomly sampling 10,000 images from a StyleGAN2 \cite{karras2020analyzing} model, only 26 corresponded to Indians, 171 to Africans, while over 6500 were Caucasians. Thus, simple rejection sampling is not only inefficient but also implausible for generating a large number of samples for underrepresented groups.

Previous methods that aimed to generate data for specific protected attributes, such as race, have either trained a generative model from scratch or fine-tuned an existing model. However, both of these approaches require the collection of large amounts of real data for the target-protected attribute. For instance, a recent study \cite{sevastopolsky2022boost} collected over 5 million images of Africans and 3 million images of Asians from various YouTube sources.

In contrast, we propose a simple and novel search-based algorithm that exploits the significant variability that can be derived by the use of existing generative models. Previous research has shown that different facial or protected attributes correspond to latent directions or high-dimensional latent spaces \cite{colbois2021use}. Based on this fact, we develop a search-based algorithm that can generate a large number of unique synthetic identities in a zero-shot manner. We generate over 50,000 synthetic identities for six different racial groups, including Indian, White, African, Black, Asian, Middle Eastern, and Latino Hispanic, resulting in a total of 13.5 million images with 45 images per person.

Finally, we demonstrate that pretraining on this large corpus of balanced data can significantly improve the performance of facial recognition systems. We evaluate three different systems, namely ArcFace \cite{deng2019arcface}, AdaFace \cite{kim2022adaface}, and ElasticFace \cite{boutros2022elasticface}, and show that our approach outperforms models trained using widely accepted unbalanced datasets such as VGGFace2 \cite{cao2018vggface2}. Even the BUPT balanced-face dataset \cite{wang2020mitigating}, which already contains an equal number of identities per race, shows improvements with our approach.

The proposed approach of generating a balanced dataset from a biased generative model can also be useful for other downstream tasks, such as ethnicity or age classification, and not just limited to facial recognition. Additionally, the approach can be applied to other protected attributes besides race, such as age and gender.

We make the following contributions in this paper:
\begin{itemize}
    \item We propose a simple evolutionary search-based approach to generate a large balanced set of images using an existing biased generative model. Our approach doesn't require any training or fine-tuning of the generative model. 
    \item We contribute a dataset of over 50,000 distinct synthetic identities for six different racial groups resulting in a total of 13.5 million images with 45 images per person.
    \item We show that the generated dataset improves the performance of three different facial recognition systems, namely ArcFace, AdaFace, and ElasticFace, on both balanced and unbalanced datasets.
    \item We highlight the potential of this approach to generate balanced datasets for other protected attributes and downstream tasks such as ethnicity or age classification. 
    \item We contribute to the ongoing discussion of addressing bias in facial recognition by providing a practical and scalable solution that can be implemented without requiring the collection of large amounts of real data.

\end{itemize}

\section{Related Work}
This section provides a brief overview of the most relevant work done on bias mitigation in facial recognition models, including the use of synthetic data generation to fine-tune these models, which is particularly pertinent to the proposed approach.

\subsection{Bias Mitigation in Facial Recognition}

Past research has extensively shown that widely accepted deep learning-based facial recognition algorithms exhibit biases towards a particular ethnicity \cite{singh2022anatomizing, leslie2020understanding, anastasi2006evidence, mittal2023bias}. Most of the research done on bias mitigation has been directed towards mitigating the bias for particular demographic subgroups rather than arriving at generalizable solutions across demographic groups \cite{singh2022anatomizing}. 

Zhang et al. \cite{zhang2018mitigating} proposed an adversarial learning approach to reduce bias in facial recognition systems. Yucer et al. \cite{yucer2020exploring} proposed an approach to alter the ethnicity of a person through an adversarial training procedure applied to a cyclegan model. Gong et al. \cite{gong2021mitigating} used adaptive convolutional kernels and attention mechanisms based on the demographic subgroup to mitigate demographic biases. Wang et al. \cite{wang2020mitigating} propose a reinforcement learning-based race balance network. They also introduce the BUPT-GlobalFace and BUPT-BalancedFace datasets containing datasets with racial distribution on global and balanced distributions respectively. Researchers have also proposed approaches to remove protected attributes in data representations by adversarial training models \cite{wang2019balanced, kim2019learning, zhang2018mitigating}.

Other researchers who studied biases in facial recognition models have pointed out correlations between protected attributes and other features which is often the reason for biases in face recognition models. Researchers have proposed approaches to disentangle the protected attribute from other features \cite{park2021learning} as well as suppressing the protected attribute \cite{dhar2021pass}
to have fairer face recognition models. In our case, given the control that generative models provide, we can ensure that some attributes such as pose, illumination, and expression do not correlate with the protected attribute.

\begin{figure*}[tb]
\centering
\subfloat[Seed]{\includegraphics[width=0.12\paperwidth]{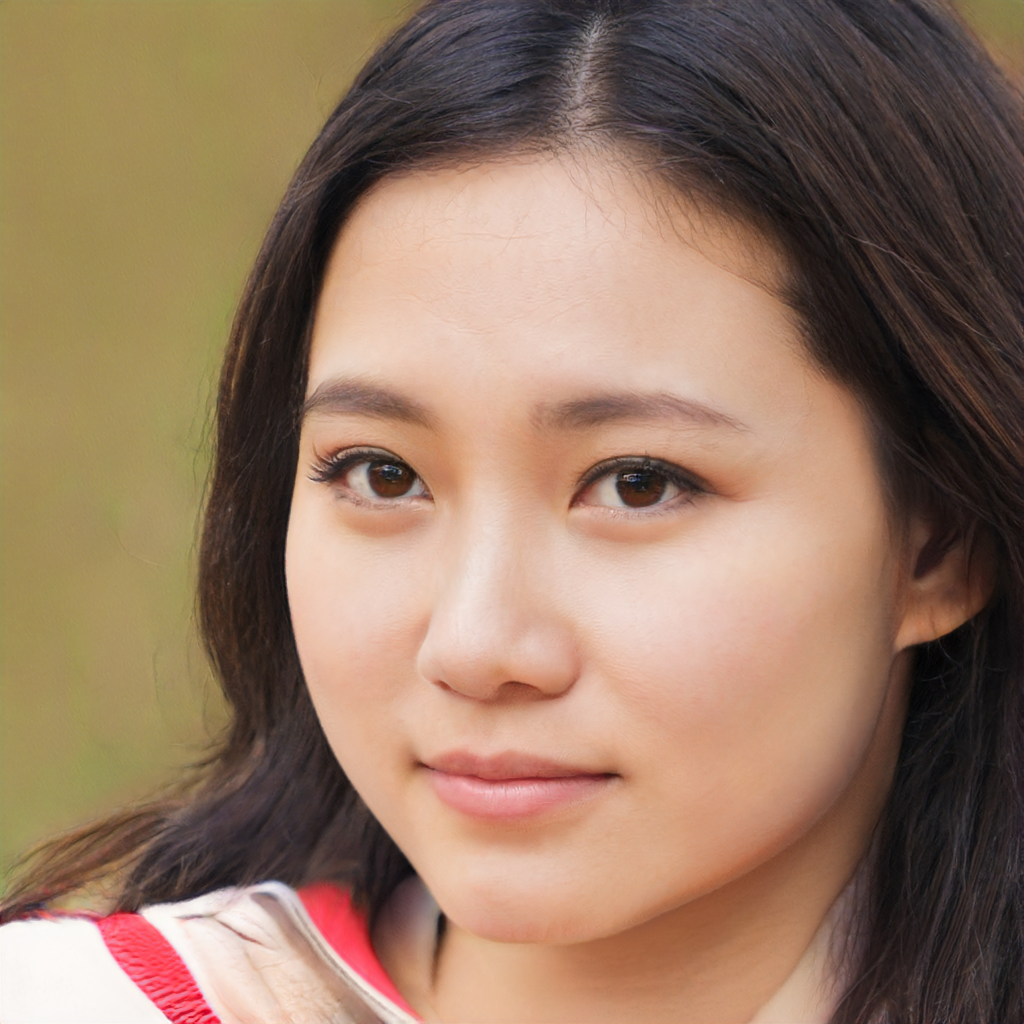}}
~
\subfloat[Iteration 1]{\includegraphics[width=0.12\paperwidth]{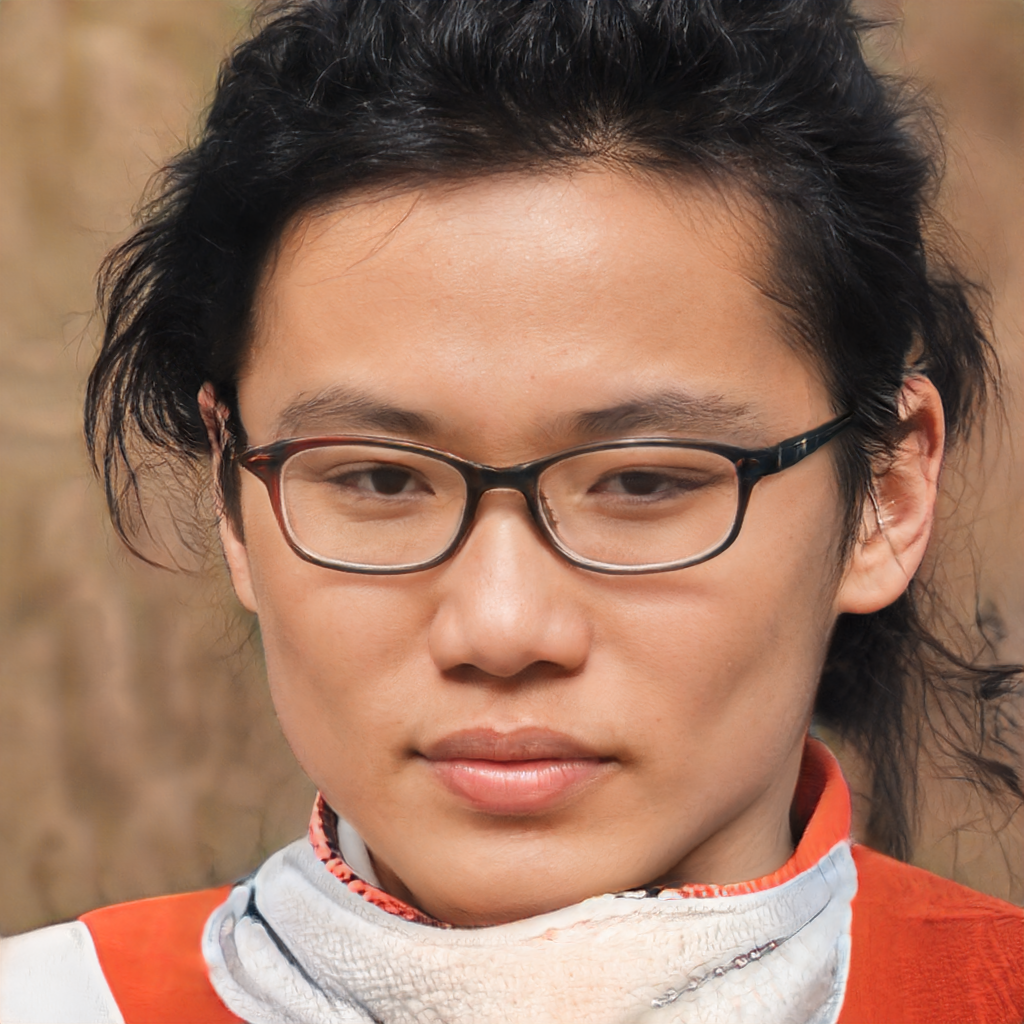}}
~
\subfloat[Iteration 2]{\includegraphics[width=0.12\paperwidth]{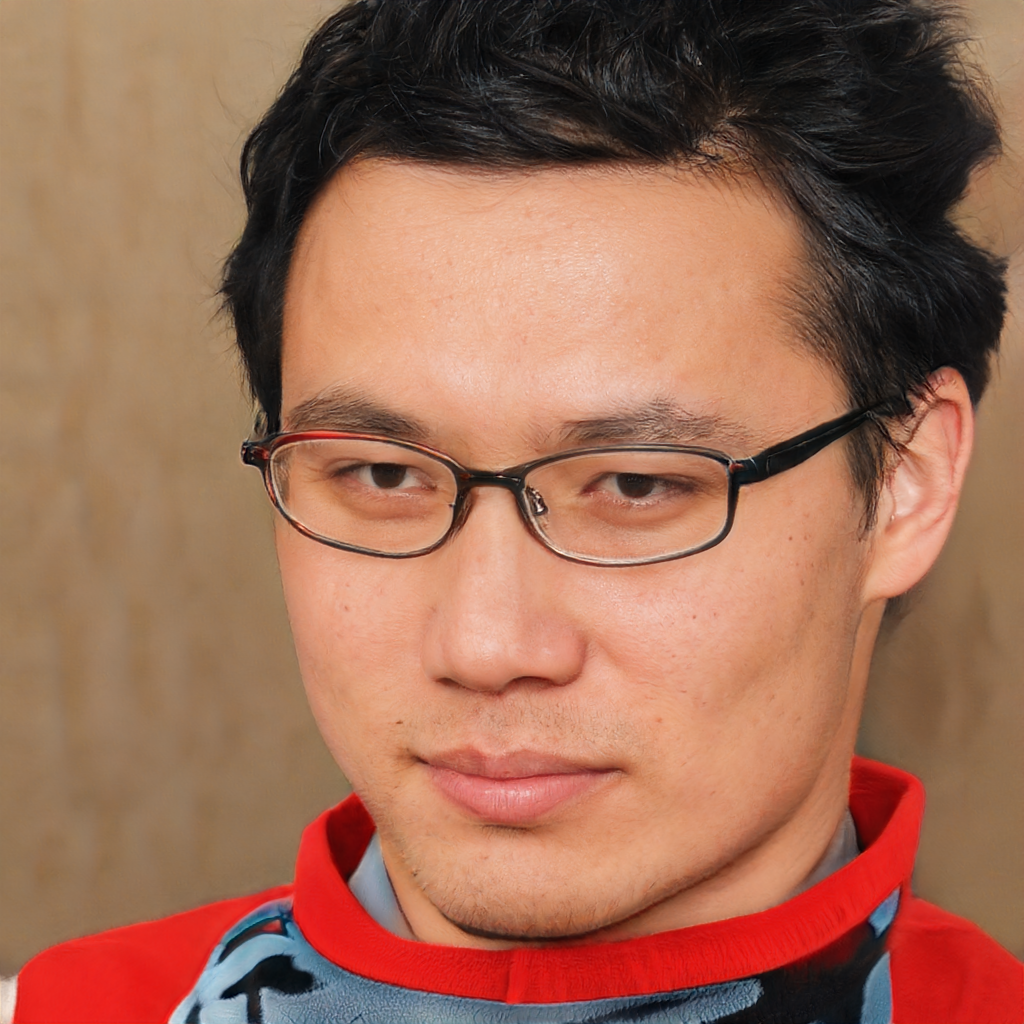}}
~
\subfloat[Iteration 3]{\includegraphics[width=0.12\paperwidth]{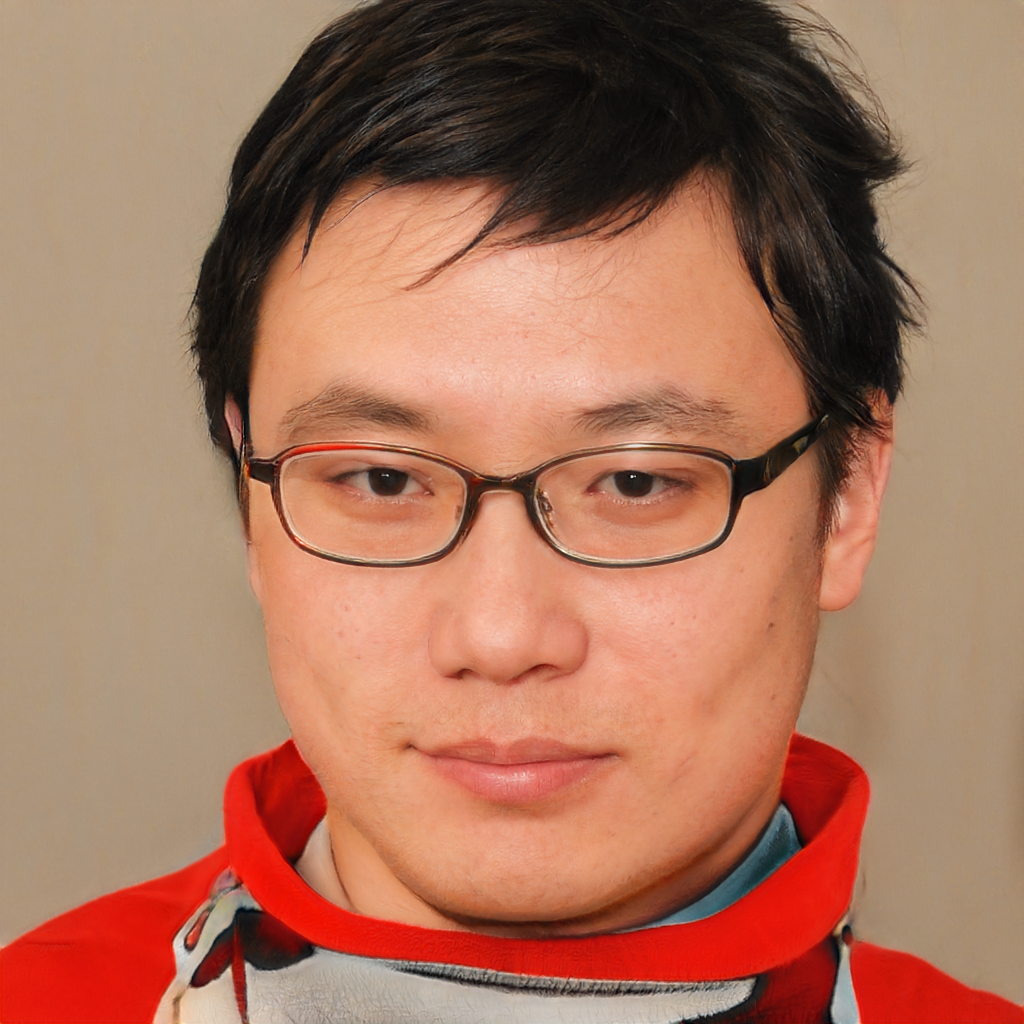}}
~
\subfloat[Iteration 4]{\includegraphics[width=0.12\paperwidth]{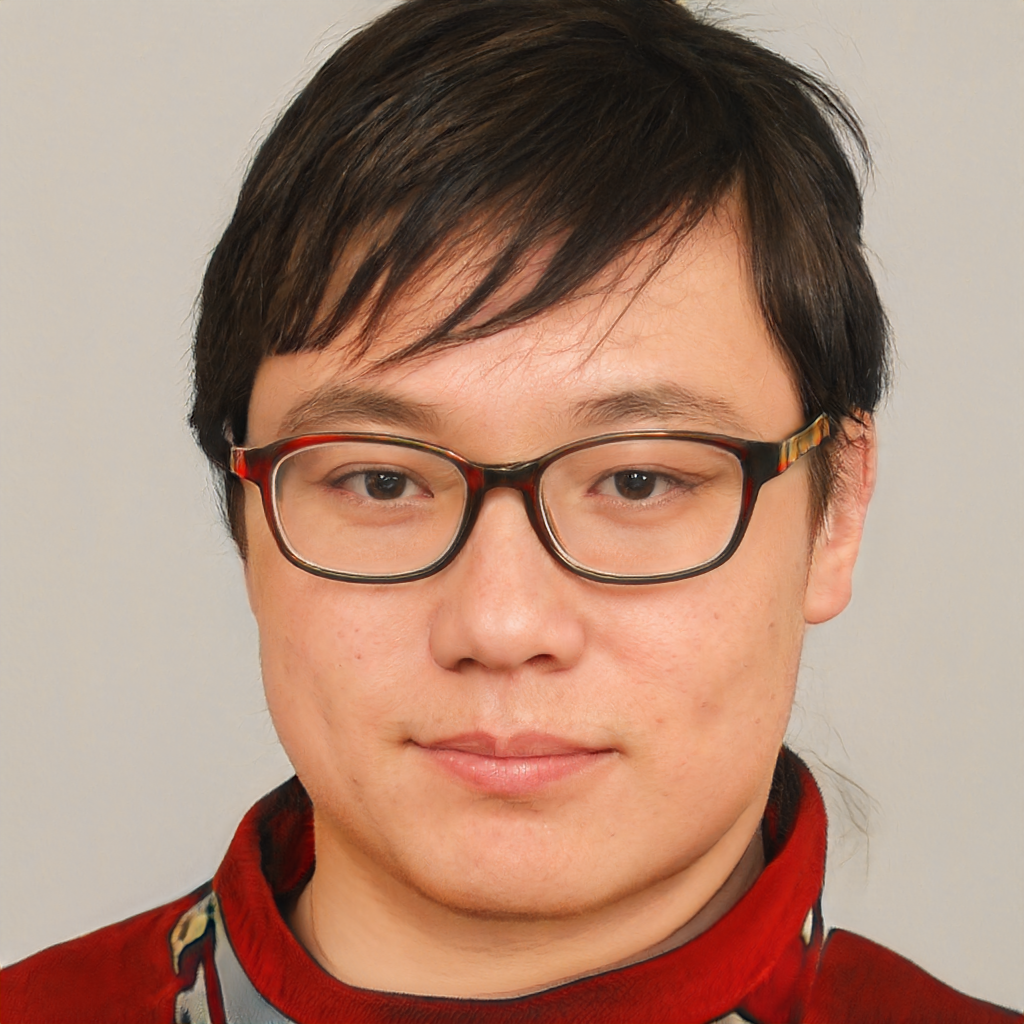}}
~
\subfloat[Iteration 5]{\includegraphics[width=0.12\paperwidth]{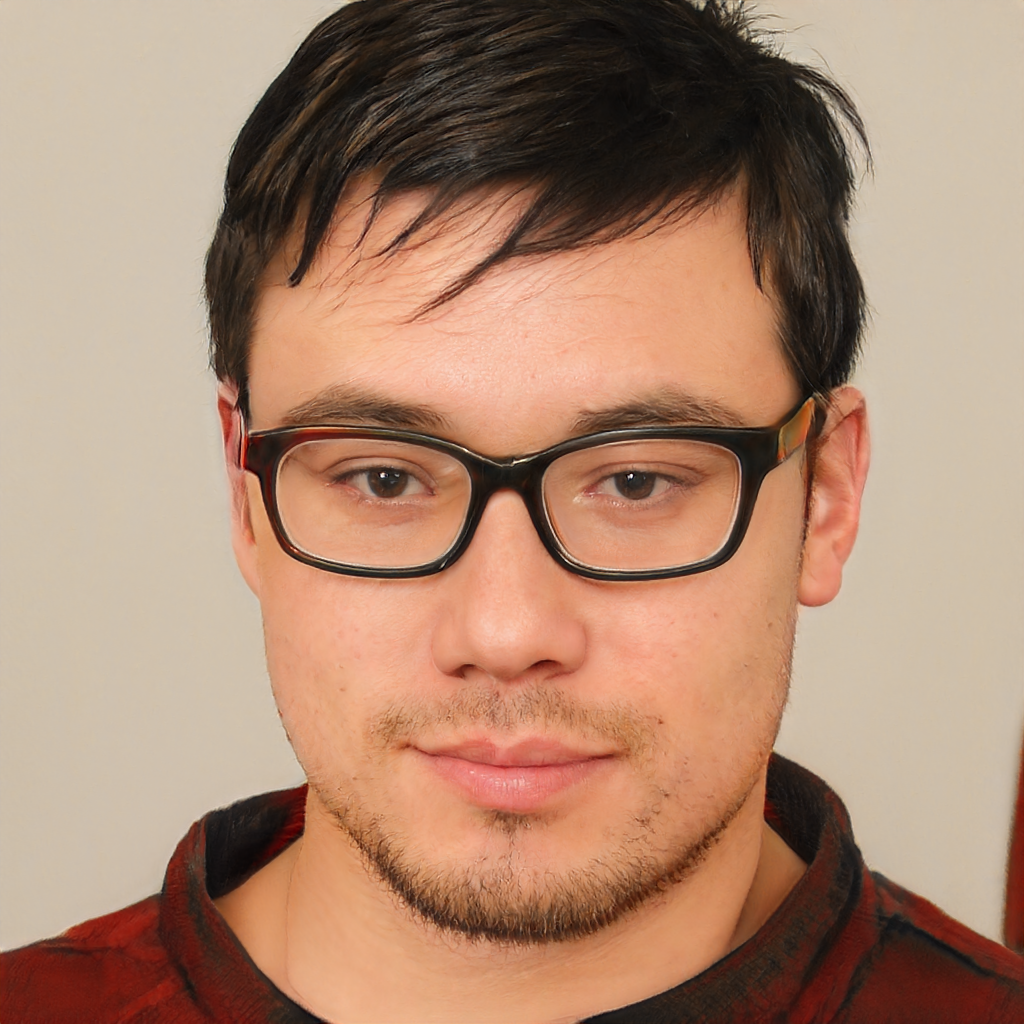}}
\\
  
\caption{Images showing the generation of synthetic data using our approach targeting the 'Asian' ethnic group while mutating in the z-subspace with the number of mutations set to 1.}
\label{fig:iterations}
\end{figure*}

\subsection{Synthetic Data for Face Recognition Systems}

Researchers have shown interest in the use of synthetic data for face recognition systems due to their privacy-preserving properties. Most researchers have trained new generative models on different types of real datasets \cite{kortylewski2019analyzing, boutros2022sface, sevastopolsky2022boost, kim2023dcface}. Boutros et al. \cite{boutros2022sface} proposed an approach to train facial recognition using synthetic images. They trained a generative model conditioned on the user identity to create a synthetic dataset. Sevastopolsky et al. \cite{sevastopolsky2022boost} also proposed an approach to train a generative model on unlabelled data collected from Youtube. They use this model to train an encoder model which is then finetuned for face recognition. 

However, all of these methods require a large number of either labeled or unlabelled images for training the generative models for their task. In this paper, we eliminate this step by making use of an  existing generative model even though it is biased. In addition to using a large number of real data, these approaches either don't consider demographic groups or have specifically targeted particular subgroups. They are not generalizable to other protected attributes or subgroups. Thus for these reasons, we opt to use existing trained generative models on a generic real-world biased dataset and propose an approach that is independent of the demographic subgroup or category.

A more related line of research to our work has been on using pre-trained generative models to edit images by traversing the latent space. Colbois et al. \cite{colbois2021use} demonstrated that specific latent directions exist in the StyleGAN2 latent space that can modify the pose, illumination, and expressions of synthetic identity. Similarly, Ramaswamy et al. \cite{ramaswamy2021fair} proposed a method to de-correlate target labels (e.g. glasses, hats) with protected attributes (e.g. race, gender) to remove biases in facial attribute classification models. Other researchers have also proposed approaches to alter the facial attributes in images \cite{dabouei2020boosting, shen2020interpreting, he2019attgan, liu2019stgan, parmar2022spatially}. Researchers have also found ways to disentangle the identity of a person from other attributes of the image \cite{le2022styleid, nitzan2020face}. However, due to the limitations of these approaches, they are still not adequate for training facial recognition models. Thus, to the best of our knowledge, no one has proposed a method for generating new identities with specific demographic characteristics while using a pre-trained biased generative model.

\section{Proposed Approach}

In this section, we present our approach to training fair facial recognition models using synthetically generated data. First, we discuss the approach to generating racially balanced synthetic identities in section \ref{sec:generate}. Then, in section \ref{sec:train}, we explain how we use this data to train facial recognition models.

\subsection{Generating demographic-specific identities}
\label{sec:generate}

Given the biased nature of existing generative models they cannot directly be used for training downstream tasks on synthetic data as they would propagate similar biases. If the generative model was unbiased, an efficient technique to generate a balanced dataset can be by random rejection sampling. However, in reality, given the biased nature of the generate models and the StyleGAN models in particular, it is extremely time-consuming for sampling identities belonging to the underrepresented groups (approximately 14 hours to generate just 1000 "Black" individuals, refer to section \ref{sec:time_complexity}). We propose using a controllable latent space search algorithm that can address this problem without needing to re-train or finetune these models on any real data. We use a pre-trained StyleGAN2\cite{karras2020analyzing} model that was trained on the Flickr-Faces High-Quality \cite{karras2019style} dataset to generate 1024x1024 closeup facial images. The StyleGAN2 model uses two latent spaces, the initial Gaussian random space $Z \in \mathbb{R}^{512}$  which is mapped to a larger latent space $W \in \mathbb{R}^{18\times512}$ using a small neural network-based mapping function.

Previous research has shown the existence of subspaces in the StyleGAN latent space pertaining to individual demographic groups. These can either be explored by latent directions for attributes such as lighting, expression and pose \cite{colbois2021use}. However, unique identities for different protected attributes cannot be generated using latent directions as the identity and attribute are correlated, that is, you cannot change the identity and the attribute simultaneously. In section \ref{sec:discussion}, we have discussed in detail issues with adapting previously proposed approaches in editing GAN images to this task. Our main considerations are for the proposed approach to be privacy-aware and thus not utilizing any real data. We have also taken into consideration the memory and computational time complexity.

Taking these factors into consideration, we propose using a controllable latent space search algorithm similar to a breadth-first search on the latent space of a StyleGAN2 model. We start randomly sampling latent vectors till we find one that when passed through the generator generates an image corresponding to the target attribute. We use an auxiliary demographic classifier \cite{serengil2020lightface} for checking whether the generated image belongs to the target group. Using this demographic classifier we define the fitness function as follows, 

\begin{equation}
f(\textbf{v})= \begin{cases}1 & if \ \  C(G(v)) = T\\0 & otherwise\end{cases}
\end{equation}

where T is the target demographic subgroup, G is the generative model, and C is the demography classifier. We use the latent vector found through random sampling as the starting point, referred to as $\textbf{v}_s$ in the pseudo-code \ref{algo:pseudo-code}. As is the case with breadth first search we maintain a queue for the traversal. The starting point is added to a queue and we use this to begin the search. 

Iteratively, we dequeue a latent vector from the queue, referred to as $\textbf{v}_c$. If $f(\textbf{v}_c)=1$ then, we sample neighboring points by mutating the current latent vector $\textbf{v}_c$ with a random uniform variable. We use a uniform random variable instead of a multivariate Gaussian random variable, which is typical in most search algorithms, as it provides better control in the GAN latent space and allows us to generate reasonable facial images by staying within appropriate boundaries. Let $\textbf{v}_i$ be the set of vectors mutated from the vector $\textbf{v}_c$. 

\begin{equation}
    \textbf{v}_i = \{ \textbf{v}_{i1}, \textbf{v}_{i2}, ..., \textbf{v}_{in} \}
\end{equation}

where $n$ is the number of mutations of the current vector $\textbf{v}_c$. We append the entire set $\textbf{v}_i$ into the queue. We continue the search process till the queue is not empty and other search controls such as maximum iterations haven't been exceeded. We show the iterative generation of latent identities in Figure \ref{fig:iterations} for the Asian demographic subgroup with the number of mutations set to 1.

%This allows for a controlled evolution of the current vector.   

% To start, we randomly sample latent vectors until we find a vector X (which could either belong to the Z or W+space), that corresponds to the demographic group we are targeting. Then, we evolve the vector X using a random uniform variable that always moves further away from the starting vector. Using a uniform random variable instead of a multivariate Gaussian random variable, which is typical in most search algorithms, provides better control in the GAN latent space and allows us to generate reasonable facial images while staying within appropriate boundaries. We use an auxiliary demographic classifier \cite{serengil2020lightface} as feedback for the evolution, only exploring directions that support the target demographic. Additionally, we use a Google mediapipe \cite{lugaresi2019mediapipe} face detection model to stay within the bounds of the W+ latent space. This is however not required when evolving the Z latent space. Thus the search algorithm continues to progressively mutate away from the starting point till it either hits the bounds of the space or the demographic group. We also see that there are advantages of using multiple seed values as the latent space is extremely high dimensional and it is easy to go out of bounds in any one particular dimension. \ref{pseudo-code} provides the pseudo-code for the proposed algorithm. 

Our approach works well on both the W and Z latent vector space, even though previous research has suggested limited disentanglement of the Z space. The trade-off here is similar to the use of a truncation-psi parameter, between the diversity and quality of the images. The Z latent space ensures better quality, however with limited diversity. In contrast, the W latent space ensures higher diversity but may compromise the quality of the image. This also requires us to use a Google mediapipe \cite{lugaresi2019mediapipe} face detection model to stay within the bounds of the W latent space. This is however not required when evolving the Z latent space. Also, in the case of mutating in the W-latent vector space, we observed that after a large number of iterations ($> 500$), when a number of possible directions had been exhausted, the synthetic identities start looking similar. This is because the model is forced to take directions where the identity doesn't change but only variants of the same identity are produced. We see that limiting the number of iterations from a particular seed value can efficiently take care of this problem.  

The proposed approach runs independently for different ethnicities as shown in the pseudo-code \ref{algo:pseudo-code} and can thus parallelly generate data for different ethnicities. This allows us to specifically control how many samples are required for each demographic subgroup and we can appropriately terminate the search operation once this criterion is satisfied.

%We demonstrate that this approach can still find synthetic identities corresponding to different demographic groups, albeit with lesser diversity. But it allows us to generate images with looser constraints are the mapping network ensures that we do not traverse out of bounds of generating reasonable face images. We still recommend the $W$ space due to the reasons discussed earlier. 

For generating multiple images for each identity, we use the approach proposed by Colbois et al. \cite{colbois2021use} using latent directions for generating expression, pose, and illumination variations while preserving the identity. We have, however, excluded extreme pose variations due to the limitations of the approach on StyleGAN2 as shown by the original authors. This allows to specifically curate a diverse range of facial expressions, poses and illumination while maintaining consistency across identities and ethnicities. 

In this study, we have broadly classified images into 6 ethnic groups using an auxiliary ethnicity classifier \cite{serengil2020lightface} - Caucasian, African, Indian, Asian, Middle Eastern, and Latino Hispanic. In comparison to previous studies that have generally used only 4 racial groups, we believe this is more inclusive even though the test face recognition datasets only contain labels for 4 groups - Indian, African, Caucasian, and Asian. We generate two versions each for the Z and W spaces containing 15,000 and 50,000 synthetic identities per race with 25 and 45 images per person respectively. These are referred to as z-15k and z-50k for the Z latent space and w-15k and w-50k for the W latent space. 

%We generate 50,000 synthetic identities for each demographic group and 45 images per person. In total, we generate around 13.5 million images. 

\begin{algorithm}[t]

   \caption{Latent space exploration for generating synthetic identities for each demographic subgroup.}
   % \label{algo:DeepReShape}
   \label{algo:pseudo-code}
   \textbf{Input:} A generative model $G$, an auxiliary ethnicity classifier $C$,  target race $t$, starting latent vector found using random sampling $v_s$ s.t. $C(G(v_s)) = t$, number of mutations $n$, max range of random mutation $\delta$, maximum number of iteration for a particular starting vector $max\_iter$ \\
   \textbf{Output:} A list of latent space vectors w. 
   
   \begin{algorithmic} [1]
   % \State $seed \leftarrow s$
   \State $queue \leftarrow []$ %\Comment{Initialize empty list}
   \State $out \leftarrow []$ %\Comment{Initialize empty list}
    \State $iter \leftarrow 0$
    \State $queue.enqueue(v_s)$ 

   \While{ $len(queue) \neq 0$ and $iter \leq max\_iter$ }
   
   \State $current\_vector = queue.dequeue()$
   % \For{$i$ = 1 {\bfseries to} n} 
    \State $image = G(v_s)$
   \If{ face not detected in image}
   \State \textbf{continue}
    \EndIf
   \If{$C(image) == t$}

    \State $out.append(v_c)$ 
    \State $iter \leftarrow iter + 1$
    \For{j = 0 {\bfseries to} n}
        \State $v_j \leftarrow v_c + random(G.latent\_dims, range=[-\delta, \delta])$
        \If{$dist(v_j, v_s) > dist(v_c, v_s)  $}
            \State $queue.enqueue(v_j)$
        \EndIf
    \EndFor
   \EndIf
   \EndWhile 

   \State \textbf{return} List of latent vectors corresponding to the target ethnicity - $out$. 
   
\end{algorithmic}
\end{algorithm}

\subsection{Training the Face Recognition Model}
\label{sec:train}

 To show the advantages of this generated dataset, we pre-train face recognition models on the generated dataset. We make use of three facial recognition models, namely ArcFace, ElasticFace, and AdaFace. As a baseline, these models have been trained on just real data. We have specifically selected three popularly used datasets in face recognition with varied biases - VGGFace2, BUPT-BalancedFace, and BUPT-GlobalFace. We have followed the same training and testing protocols as the original authors of the respective recognition models. 
 
 We report results on the Racial Faces in the Wild (RFW) dataset \cite{wang2019racial} which contains partitions for 4 racial groups - Caucasians, Blacks, Indians, and Asians. We also report results on the Labeled Faces in the Wild (LFW)\cite{huang2008labeled}, Celebrities in Frontal-Profile in the Wild (CFP-FP and CFP-FF) \cite{cfp-paper}, AgeDB \cite{moschoglou2017agedb}, Cross-Age LFW (CALFW) \cite{calfw} and the Cross-Pose LFW (CPLFW) \cite{CPLFWTech} datasets. We have provided a detailed description of all the training and testing datasets in the Appendix section \ref{sec:dataset_details}. We have provided the training and fine-tuning details including the hyperparameters in the Appendix \ref{sec:hyperparameters}.

\begin{table}[htbp]
  \centering
\small
% \fontsize{8pt}{8pt}\selectfont
\addtolength{\tabcolsep}{-2pt}
  \caption{Recognition performance when training on the VGGFace2 dataset and testing on the subsets of the RFW dataset. The models trained on real data serve as a baseline. }
    \label{tab:rfw-vgg}
\begin{tabular}{ccccccc}
\hline
Model & Dataset & & RFW & & & AD \\
& & Indian & Asian & White & African & \\ \hline 
ArcFace & Real & 79.17 & 74.90 & 82.48 & 73.80 & 8.68 \\ \cline{2-7}
& w-15k & 79.43 & 75.42 & 82.03 & 74.58 & 7.45 \\
& z-15k & 77.83 & 74.77 & 81.53 & 74.40 & 7.13 \\
& w-50k & \textbf{80.58} & \textbf{77.10} & \textbf{83.12} & \textbf{76.47} & \textbf{6.65} \\
& z-50k & 80.28 & 76.92 & \textbf{83.12} & 75.95 & 7.17 \\ \hline
Adaface & Real & 77.97 & 75.67 & 82.52 & 70.18 & 12.33 \\ \cline{2-7}
& w-15k & 77.53 & 74.50 & 80.88 & 70.15 & 10.73 \\
& z-15k & 78.17 & 75.38 & 81.30 & 71.72 & \textbf{9.58} \\
& w-50k & 77.20 & 76.07 & 82.02 & 68.40 & 13.62 \\
& z-50k & \textbf{79.15} & \textbf{77.55} & \textbf{82.73} & \textbf{72.48} & 10.25 \\ \hline
Elasticface & Real & 74.97 & 71.32 & 77.78 & 70.92 & \textbf{6.87} \\ \cline{2-7}
& w-15k & 79.78 & 75.87 & 83.47 & 75.90 & 7.57 \\
& z-15k & \textbf{81.10} & 76.23 & \textbf{84.62} & \textbf{77.08} & 7.53 \\
& w-50k & 80.92	 & \textbf{76.88} &	83.83 &	76.32&	7.52 \\
& z-50k & 79.90 & 75.52 & 83.72 & 75.63 & 8.08 \\ \hline 
\end{tabular}
\end{table}

\begin{table}[htbp]
\centering
\small 
\addtolength{\tabcolsep}{-2pt}
\caption{Recognition performance when training on the BUPT-BalancedFace dataset and testing on the subsets of the RFW dataset. }
\label{tab:rfw-bupt}
\begin{tabular}{ccccccc}
\hline 
Model & Dataset & & RFW & & & AD \\
& & Indian & Asian & White & African & \\ \hline 
ArcFace & Real & 94.23 & 92.87 & 95.03 & 92.92 & 2.12 \\  \cline{2-7}
& w-15k & \textbf{94.97} & \textbf{93.70} & \textbf{95.35} & 93.15 & 2.20 \\ 
& z-15k & \textbf{94.97} & 93.67 & 95.25 & \textbf{93.67} & \textbf{1.58} \\ 
& w-50k & 93.73 & 92.75 & 94.95 & 92.38 & 2.57 \\ 
& z-50k & 94.43 & 93.00 & 94.77 & 92.48 & 2.28 \\ \hline
Adaface & Real & 93.28 & 92.87 & 95.02 & \textbf{90.78} & 4.23 \\ \cline{2-7}
& w-15k & \textbf{93.43} & 92.87 & \textbf{94.97} & 90.23 & 4.73 \\ 
& z-15k & \textbf{93.43} & \textbf{92.97} & 94.62 & 89.80 & 4.82 \\ 
& w-50k & 93.33 & 92.30 & 94.22 & 90.18 & \textbf{4.03} \\ 
& z-50k & 93.38 & 92.57 & 94.37 & 89.92 & 4.45 \\ \hline 
Elasticface & Real & 94.23 & 93.83 & 95.30 & 93.03 & 2.27 \\  \cline{2-7}
& w-15k & 94.55 & \textbf{93.98} & 95.68 & 93.15 & 2.53 \\ 
& z-15k & \textbf{94.70} & 93.97 & \textbf{96.02} & \textbf{93.82} & 2.20 \\ 
& w-50k & 94.63 & 93.50 & 95.85 & 93.52 & 2.33 \\ 
& z-50k & 94.22 & 93.60 & 95.77 & 93.67 & \textbf{2.10} \\ \hline
\end{tabular} 
\end{table}

\begin{table*}[htbp]
\centering
\small
% \fontsize{8pt}{8pt}\selectfont
% \addtolength{\tabcolsep}{-3pt}
\caption{Recognition performance when training on the VGGFace2 dataset and testing on the LFW, CFP-FP, CFP-FF, CALFW, AgedB, and the CPLFW datasets. }
\label{tab:rest-vgg}
\begin{tabular}{cccccccc}
\hline 
Model & Train Dataset & LFW & CFP-FP & CFP-FF & AGED-DB & CALFW & CPLFW \\ \hline 
ArcFace & Real & 81.95 & 57.63 & 68.51 & 55.77 & 90.45 & 80.60 \\ \cline{2-8}
& w-15k & 82.10 & 56.90 & 67.83 & 55.82 & 90.13 & 80.70 \\
& z-15k & 81.82 & 58.00 & 68.90 & 55.55 & 90.05 & 81.33 \\
& w-50k & \textbf{82.37} & \textbf{58.96} & 69.14 & 55.55 & 90.82 & \textbf{81.50} \\
& z-50k & 81.55 & 57.59 & \textbf{69.51} & \textbf{56.17} & \textbf{91.27} & 81.00 \\ \hline 
Adaface & Real & 96.58 & 85.49 & 97.80 & \textbf{84.37} & 88.97 & 79.25 \\ \cline{2-8}
& w-15k & 95.57 & 82.89 & 97.11 & 83.33 & 88.47 & 77.85 \\
& z-15k & 95.98 & 83.82 & 97.33 & 84.30 & 89.30 & 78.47 \\
& w-50k & \textbf{96.75} & \textbf{87.03} & 97.67 & 81.63 & 88.57 & \textbf{79.72} \\
& z-50k & 96.68 & 85.87 & \textbf{98.10} & 84.22 & \textbf{89.42} & 79.52 \\ \hline 
Elasticface & Real & 81.18 & 58.99 & 67.49 & 54.40 & 87.70 & 76.90 \\ \cline{2-8}
& w-15k & 83.20 & 61.13 & 70.36 & 56.40 & 91.20 & 81.52 \\
& z-15k & \textbf{84.15} & 60.64 & 70.96 & 56.63 & 91.97 & \textbf{83.60} \\
& w-50k & 83.05	& 60.87 &	\textbf{71.10} &	\textbf{56.88} &	\textbf{92.82} &	82.60 \\
& z-50k & 83.23 & \textbf{61.16} & 70.90 & 56.45 & 91.02 & 81.23 \\ \hline 
\end{tabular}

\end{table*}

\begin{table*}[htbp]
\centering
\small
% \fontsize{8pt}{8pt}\selectfont
% \addtolength{\tabcolsep}{-3pt}
\caption{Recognition performance when training on the BUPT-BalancedFace dataset and testing on the LFW, CFP-FP, CFP-FF, CALFW, AgedB, and the CPLFW datasets. }
\label{tab:rest-bupt}
\begin{tabular}{cccccccc}
\hline 
Model & Dataset & LFW & CFP-FP & CFP-FF & AGE-DB & CALFW & CPLFW \\ \hline 

ArcFace & Real & 86.48 & 60.50 & \textbf{71.26} & \textbf{58.27} & 94.77 & 90.62 \\ \cline{2-8}
& w-15k & \textbf{87.25} & 59.29 & 70.50 & 57.75 & 95.28 & 90.43 \\
& z-15k & 86.92 & \textbf{60.87} & 70.51 & 57.34 & \textbf{95.35} & \textbf{91.03} \\
& w-50k & 86.83 & 60.04 & 70.50 & 57.33 & 95.18 & 90.53 \\
& z-50k & 86.88 & 58.66 & 70.61 & 57.20 & 95.03 & 90.00 \\ \hline
Adaface & Real & 99.43 & 88.14 & 98.70 & \textbf{91.97} & \textbf{94.97} & \textbf{88.38} \\ \cline{2-8}
& w-15k & 99.40 & 87.07 & 98.63 & 91.92 & 94.83 & 87.53 \\
& z-15k & 99.25 & 87.69 & \textbf{98.97} & 90.63 & 94.65 & 87.72 \\
& w-50k & \textbf{99.43} & \textbf{88.86} & 98.67 & 91.62 & 94.73 & 88.10 \\
& z-50k & 99.25 & 87.99 & 98.50 & 90.68 & 94.53 & 87.52 \\ \hline 
Elasticface & Real & 87.52 & 61.30 & 71.60 & 58.30 & 95.13 & 91.28 \\ \cline{2-8}
& w-15k & 88.00 & 61.41 & 71.19 & 58.25 & 95.13 & \textbf{91.92} \\
& z-15k & 87.80 & \textbf{61.77} & \textbf{72.41} & 58.15 & \textbf{95.35} & 91.63 \\
& w-50k & \textbf{88.15} & 61.16 & 71.71 & \textbf{58.68} & 95.08 & 91.25 \\
& z-50k & 87.92 & 61.73 & 72.00 & 57.82 & 95.15 & 91.40 \\ \hline 

\end{tabular}
\end{table*}

\subsubsection{Metrics used for evaluation}

For evaluation of the facial recognition models, we report the recognition accuracy on various datasets discussed in the previous section that highlight the advantages of our approach in different scenarios. Additionally, similar to \cite{jain2022dataless} we utilize the recognition accuracy difference (AD), metric for evaluation of the post-training model biases. Accuracy difference is the maximum difference or disparity between the classification accuracy of different facets in a set of demographics $ \{ d_1, d_2, ..., d_n \} \in D$ (equation \ref{eq:ad}).

\begin{equation}
    AD = max_{i,j} | ACC_i - ACC_j| \forall i,j \in D
    \label{eq:ad}
\end{equation}

% \begin{equation}
%     DTPR = max_{i,j} | TPR_i - TPR_j| \forall i,j \in D
%     \label{eq:dar}
% \end{equation}

\section{Results on the use of the synthetic dataset for training facial recognition. }

We hypothesize that using a balanced dataset for different ethnicities will lead to a more accurate and fair face recognition model. By exposing the model to a diverse range of pose, gender, and racial variations, we can reduce the likelihood of biases while boosting the performance of the face recognition model. A balanced dataset will help ensure that the model is better able to recognize individuals from underrepresented communities, who may be more likely to be falsely identified by traditional biased models. Additionally, since we expect there to be a domain shift from high-quality synthetic images to real-world in-the-wild datasets, we finetune all the models trained on balanced synthetic datasets on real-world datasets.

Most facial recognition models currently are trained on datasets such as MS-1M, and VGGFace2. All of these datasets are unbalanced with respect to ethnic diversity. We compare the advantages of pretraining on the generated synthetic dataset as compared to only training a face recognition model on real data. We summarize the results on the RFW dataset in table \ref{tab:rfw-vgg} for the VGGFace2 dataset. We have included results for the BUPT-Globalface dataset in the Appendix in Table \ref{tab:rfw-global}. We see a significant improvement in the performance of the models. Especially in the case of the ElasticFace model when finetuned on the VGGFace2 dataset, the recognition accuracy on the RFW dataset improves from 74.97\% to 81.10\% for Indians, 71.31\% to 76.23\% for Asians, 77.78\% to 84.62\% for Caucasians and  70.92\% to 77.08\% for Africans. Similarly for the AdaFace model, on average, we see an improvement of approximately 2\% in the model that was pre-trained with the z-50k synthetic dataset from the one that was trained only on the VGGFace2 dataset. 

We also see improvements in the set of experiments involving the Bupt-GlobalFace dataset. However, it is important to note that the RFW dataset has been extracted from the same MS-1M celeb dataset that BUPT-GlobalFace was created from. While the two sets are disjoint, the datasets are similar in terms of the procurement procedures. Thus, the finetuning on the Bupt-GlobalFace process played a larger role in the final performance. We believe this is the reason behind the lesser improvements for these sets of experiments. Nonetheless, for the model that was pre-trained on the Z-50k synthetic dataset, we see an improvement of approximately 0.5\% on average across the different subsets of the RFW dataset. 

Finally, we see consistently better recognition accuracy for the other dataset - LFW, AgeDB, CFP-FP, CFP-FF, CPLFW, and, CALFW - for both the set of experiments when finetuning on the VGGFace2 dataset and the BUPT-GlobalFace dataset. These results have been summarized in table \ref{tab:rest-vgg} and in the appendix table \ref{tab:rest-global} for the VGGFace2 and the BUPT-GlobalFace datasets respectively. We see the maximum improvements in the case of the AdaFace model when compared to pretraining on W-50k. The performance on the CFP-FP dataset improves from  85.49\% to 87.03\%. We can attribute any improvement in performance improvements in the CFP-FP dataset to the presence of profile view images in the synthetic training set. We made use of the control over the generative process to include all 180-degree pose variations in the training set.

Finally, we compare the performance of pretraining on synthetic data with the BUPT-BalancedFace dataset which is already balanced in terms of ethnic diversity. However, we still see improvements over the normal training regime. A recent work \cite{sevastopolsky2022boost}, showed an improvement in the range of 0.45\% to 1\% for different ethnicities on the RFW dataset by using synthetic data along with the BUPT-BalancedFace dataset. We show similar improvements in the range of 0.32\% to 0.83\% for the ArcFace model pre-trained on the W-15k dataset. It is also important to note that our approach has other added advantages - we do not collect any real data and make use of an already existing generative model. We can do the synthetic data generation in a zero-shot manner using a simple search-based approach without the requirement for training the StyleGAN from scratch as done by \cite{sevastopolsky2022boost}. We summarize the results on the RFW dataset in table \ref{tab:rfw-bupt} and on the other face recognition datasets in table \ref{tab:rest-bupt}.

We also show that the models pre-trained on the balanced synthetic data help in mitigating the bias in the model. For the Adaface models trained on the VGGFace2 dataset, we see a 2.55\% reduction in the maximum disparity between different racial groups. Similarly, there is a 2.02\% reduction for the Arcface model. However, there is a slight increase in the accuracy difference for the ElasticFace model. We see similar improvements even in the case of the models trained on BUPT-BalancedFace which is already trained on a unbiased dataset. Thus, the approach boosts the performance of the models while simultaneously reducing the bias.

\begin{figure}
    \centering
    \includegraphics[width=0.9\columnwidth]{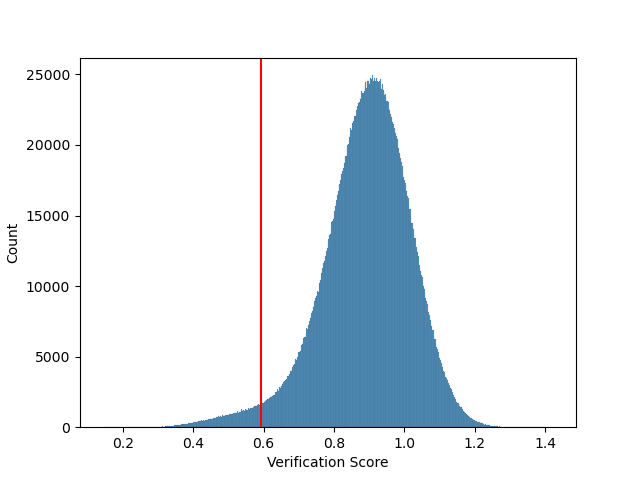}
    \caption{Histogram showing the verification scores on the generated identities using the W subspace of stylegan2. The red line depicts the operating threshold set by DeepFace. }
    \label{fig:hist_w_old_sface}
\end{figure}

% \section{Experimental Results}

% \subsection{Does the synthetic data allow efficient training FR models?}

\subsection{Are the synthetic identities biometrically different? }
\label{sec:bio_diff}

In this section, we experimentally validate that the generated identities are in-fact unique using a SOTA biometric system. The latent space search for synthetic identities is done keeping the perceptual dissimilarity between two consecutive images in mind. We do this by controlling the step size or the range of the uniform random vector. However, we don't provide the search any feedback on the biometric similarity score between two consecutive images that are generated. To validate that these images are in fact biometrically dissimilar, we perform a study using a pre-trained SFace model \cite{boutros2022sface}. The model has been taken from the DeepFace library \cite{serengil2020lightface}. We specifically choose SFace as compared to an ArcFace model as it has also been trained on synthetic data and would be better at classification on such data. It achieves similar performance on other metrics compared to the ArcFace model. 

We match each person with every other person in the dataset. We have shown the results in the form of a histogram in figure \ref{fig:hist_w_old_sface}. We do a one-to-one match for the dataset created using the W latent space. We use images with the same facial expression, pose, and illumination for every identity to remove any biases from such attributes. In the figure, the red line shows the operating threshold of 0.593 that was set by DeepFace for the SFace model. As visible in the plot, there is an extremely small tail of the histogram that is less than the threshold. Implying that the search algorithm can guarantee uniqueness with extremely high accuracy. Additionally, we do not see any low scores, which would have clearly indicated the same person being returned at multiple iterations of the search. This also implies that using this evolutionary search process we are able to generate almost $50$ times as many unique biometric identities as there were in the original FFHQ dataset for some ethnicities such as Africans and Indians. Thus, this process is not only efficient in face recognition but can also be used for other downstream tasks that require unbiased face image datasets such as ethnicity classification.

\begin{table}[]
    \centering
    \caption{Comparison of time it takes (in minutes) to generate 1000 samples through random rejection sampling versus our proposed approach for different ethnicities. We see over 200 times improvement for Indians and Blacks.}
    
\begin{tabular}{|l|l|l|}
\hline
Demographic Group& Rejection Sampling & Ours \\ \hline
Indian & 751.74 & 31.32 \\
Black & 843.84 & 17.16 \\
White & 20.62 & 13.73 \\
Asian & 118.96 & 17.28 \\
Hispanic Latino & 124.31 & 20.80 \\
Middle Eastern & 170.63 & 22.82 \\ \hline
\end{tabular}

    \label{tab:comp_time}
\end{table}

\subsection{Comparison of computational time}
\label{sec:time_complexity}

We empirically compare the computational time required for sampling racially diverse data using random rejection sampling as compared to our proposed approach. In table \ref{tab:comp_time} we show results on the time (in minutes) required for both the approaches. Random rejection sampling for under-represented groups such as Indians and Blacks is highly inefficient, and required over 12 hours to just generate 1000 samples. While our approach has comparable time for each ethnicity, requiring 32 minutes or lesser for the same number of samples. 

%This difference would further increase exponentially as the number of samples increases since a large portion of the time is required in finding the initial starting point of the algorithm. 

\section{Discussion}
\label{sec:discussion}

Although GANs offer some control over the data generation process, they also have several limitations. For instance, GANs can only change certain attributes in the variations of each identity, and they cannot replicate real-world data accurately. The generated samples are always consistent in terms of quality and size, which is not the case with real-world data. Consequently, we require additional fine-tuning on a real-world dataset to address this domain gap. 

% In the generation process we make use of an auxiliary loss 

Furthermore, while the generation process can generate perceptually different identities using various controls, they do not ensure biometric dissimilarity between different identities. A biometric model can be used as additional feedback for checking that the identities are in-fact unique. But in doing so we need to be wary of propagating the biases of the face recognition model in the generation process. Also, given the size of the dataset, this would be a costly operation to check the uniqueness with every generated identity, thus in this work, we have not used this. In other cases where uniqueness is a strict requirement this can be added as additional feedback. Furthermore, as discussed in section \ref{sec:bio_diff} we show that using a SOTA pretrained face recognition model, the synthetic identities are unique and there is a small overlap in very few cases. But again, this overlap can be attributed to the biases in the existing face recognition model. A human evaluation may be required in this regard, but it is out of the scope of this work. 

In this work, we use a state-of-the-art existing ethnicity classifier \cite{serengil2020lightface}. The approach assumes that this classifier is perfect and uses it as supervision for the evolutionary algorithm. We do not consider imperfections in the classification of the ethnicity classifier and thus we can expect some noisy predictions or misclassifications. A misclassification can occur in two different situations, the starting latent vector itself has been misclassified and the second case is where a misclassification occurs during a latent space search. Both these cases can introduce some examples of different races during the search for a particular target race but it would be limited since the classifier would need to constantly misclassify images in a particular latent subspace/ direction to continue the search there. Otherwise, the search would terminate in that latent space/ direction after its mutations return negative matches with the target race.

% {\color{red}The following para can be removed if it feels out of place?}

% We had also experimented with other approaches that are based on past research work. We have dicussed these and their limitations in Appendix \ref{sec:other_approaches}.

We had experimented with other approaches similar to past researchers \cite{colbois2021use, sevastopolsky2022boost} that project real data onto the latent space to get synthetic data. In addition to using privacy-sensitive real data, the projection approach tries to give an exact match between the real identity and projection. While for this task, we are only concerned about an estimated ethnicity match between them. This leads the projection operation to generate unclear or often even demonic faces in an effort to match other unnecessary details such as the background and clothes. Moreover, the projection operation find is more difficult for the underrepresented groups where the variations in the biased generative model are considerably lesser. Additionally, this limits the variations of synthetic data that can be generated to variations or interpolations of the projection of the real data. This would also limit the uniqueness of the identities. Along similar lines, we also experimented with randomly generating data and using these as references for these approaches instead of projecting real images on the latent space. However, due to the highly biased nature of StyleGAN2, even after generating over 100,000 samples we had very few samples for the under-represented ethnicities (<2000). This made it computationally expensive in terms of both the time required and storage space. Our proposed approach even without making use of any real or synthetic training data is able to generate a more diverse set of unique identities. This makes it both efficient in terms of time and space as it requires no training data to learn latent directions or interpolations of the data. 

It is also important to note the ethical considerations of such approaches to generate ethnicity specific data as it can be used for isolation of various groups for malicious intentions. In the work, our focus is to allow training of fairer models by creating balanced datasets and thus removing demographic biases. 

We have made our code base\footnote{\href{https://github.com/anubhav1997/youneednodataset}{https://github.com/anubhav1997/youneednodataset}} public to allow researchers to reproduce our work. Since the size of all the generated datasets combined is around 33TB, we are not able to share it directly but have made scripts available to generate them.

\section{Conclusion}

In conclusion, this work presents an approach to generate a balanced number of distinct synthetic identities for different racial subgroups from a highly biased generative model. We do so in a zero-shot manner without training or finetuning a generative model. We show that this approach works well on the stylegan2, and is successful in generating over 50,000 synthetic identities per race. Finally, we show that pretraining a face recognition model in this dataset boosts the performance of the model. Being a balanced dataset it also assists in mitigating the biases in the model and achieves fairer performance across different racial groups. We believe that this approach of generating synthetic identities is generalizable and can be used for various other downstream tasks such as age, ethnicity, gender, and, emotion classification. Our future work will be directed towards these and other applications of this work.

{\small
\bibliographystyle{plain}
\bibliography{biblio}
}

% \appendix

\section{Appendix}

\begin{table*}[htp]
\centering
\small
% \fontsize{8pt}{8pt}\selectfont
% \addtolength{\tabcolsep}{-3pt}
\caption{Recognition performance when training on the BUPT-GlobalFace dataset and testing on the LFW, CFP-FP, CFP-FF, CALFW, AgedB, and the CPLFW datasets. }
\label{tab:rest-global}
\begin{tabular}{cccccccc}
\hline 
Model & Train Dataset & LFW & CFP-FP & CFP-FF & AGED-DB & CALFW & CPLFW \\ \hline 
ArcFace & Real & 87.72 & 59.81 & \textbf{71.36} & \textbf{59.25} & 95.38 & 90.43 \\ \cline{2-8}
& w-15k & 87.58 & 58.60 & 70.66 & 58.82 & 95.62 & 90.50 \\
& z-15k & \textbf{87.82} & \textbf{60.79} & 70.87 & 59.08 & 95.40 & \textbf{91.28} \\
& w-50k & 87.27 & 57.51 & 70.86 & 58.27 & \textbf{95.63} & 89.52 \\
& z-50k & 87.28 & 58.13 & 70.87 & 58.28 & 95.62 & 89.65 \\ \hline 
Adaface & Real & \textbf{99.60} & 86.23 & 98.97 & 93.55 & 95.40 & 87.40 \\ \cline{2-8}
& w-15k & 99.57 & 85.50 & \textbf{99.16} & \textbf{93.78} & 95.13 & 87.45 \\
& z-15k & 99.38 & 85.61 & 99.00 & 93.62 & \textbf{95.63} & 87.52 \\
& w-50k & 99.50 & \textbf{86.84} & 99.07 & 93.77 & 95.30 & \textbf{87.85} \\
& z-50k & 99.47 & 85.31 & 99.10 & 93.27 & 95.48 & 87.52 \\ \hline 
Elasticface & Real & 88.80 & \textbf{61.81} & 72.20 & \textbf{60.10} & 95.58 & 91.73 \\ \cline{2-8}
& w-15k & 88.83 & 61.21 & 72.34 & 59.42 & 95.57 & \textbf{92.10} \\
& z-15k & 88.97 & 61.36 & \textbf{72.71} & 59.75 & 95.42 & 91.95 \\
& w-50k & 87.75 & 59.61 & 72.07 & 58.90 & \textbf{95.65} & 90.55 \\
& z-50k & \textbf{89.05} & 61.80 & 71.80 & 59.57 & 95.48 & 91.40 \\ \hline 
\end{tabular}
\end{table*}

\begin{table}[htp]
\centering
\small
% \fontsize{8pt}{8pt}\selectfont
\addtolength{\tabcolsep}{-2pt}
\caption{Recognition performance when training on the BUPT-GlobalFace dataset and testing on the subsets of the RFW dataset.}
\label{tab:rfw-global}
\begin{tabular}{ccccccc}
\hline
Model & Dataset & & RFW  & & & AD \\
& & Indian & Asian & White & African & \\ \hline
ArcFace & Real & 94.85 & 94.28 & 96.23 & 93.20 & \textbf{3.03} \\ \cline{2-7}
& w-15k & 95.10 & 94.65 & \textbf{97.27} & 92.97 & 4.30 \\
& z-15k & \textbf{95.33} & \textbf{95.05} & 97.00 & \textbf{93.60} & 3.40 \\
& w-50k & 94.98 & 93.78 & 96.33 & 92.60 & 3.73 \\
& z-50k & 95.17 & 94.13 & 97.10 & 92.30 & 4.80 \\ \hline
Adaface & Real & 94.22 & 93.88 & 96.63 & 91.05 & 5.58 \\ \cline{2-7}
& w-15k & 94.68 & \textbf{94.22} & 96.87 & 91.52 & 5.35 \\
& z-15k & \textbf{94.80} & 94.00 & 96.57 & \textbf{91.55} & \textbf{5.02} \\
& w-50k & 94.75 & 93.92 & 96.72 & 91.20 & 5.52 \\
& z-50k & 94.72 & 94.15 & \textbf{96.97} & 91.25 & 5.72 \\ \hline
Elasticface & Real & 95.32 & \textbf{94.70} & 97.07 & 93.68 & 3.38 \\ \cline{2-7}
& w-15k & 95.52 & 94.60 & 97.28 & 93.37 & 3.92 \\
& z-15k & \textbf{95.73} & 94.38 & \textbf{97.63} & \textbf{93.73} & 3.90 \\
& w-50k & 94.42 & 93.28 & 96.43 & 91.77 & 4.67 \\
& z-50k & 95.47 & 94.20 & 97.03 & 93.68 & \textbf{3.35} \\ \hline
\end{tabular}

\end{table}

\subsection{Details of the datasets used for training and testing FR}
\label{sec:dataset_details}

The VGGFace2 dataset contains 9000+ identities and over 3.3 million images. They do not provide any information on the numbers corresponding to each demographic group. 

The BUPT-BalancedFace contains 7000 identities per race but with small variations in the total number of images per race. The subset of the dataset with images of Caucasians contains 326 thousand images and in contrast, the subset for Indians only contains 275 thousand images. Thus even though the dataset is balanced in terms of the number of identities, it has a notable difference in the number of images per identity for the Indian subset. 

The BUPT-GlobalFace dataset mimics the demographic distribution that is prevalent in the world. It contains 38 thousand identities and 2 million images in total. 38\% percent of the dataset corresponds to white people, 31\% to Asians and 18\% to Indians, and the remaining 13\% to Africans. 

The LFW dataset contains 13,233 images of 5,749 people that were extracted using the Viola-Jones face detector algorithm. It is often referred to as the de facto benchmark for unconstrained face recognition. The CFP dataset contains images of celebrities in frontal and profile views. It contains a total of 7,000 pairs of celebrities in both the frontal-frontal (CFP-FF) and frontal-profile (CFP-FP) views. The dataset is primarily used for benchmarking the performance of face recognition across poses. The AgeDB dataset contains 12,240 images of famous personalities, including actors, writers, scientists, and politicians. It contains 440 subjects with varied ages and poses. The dataset is a good benchmark for age-invariant face recognition and age progression. It subject's ages vary from 3 years to 101 years. Similarly, the Cross-Age LFW dataset has been used as a testbed for face recognition across age groups. It has been created from the LFW dataset where 3,000 pairs of images have been selected with age gaps to add aging progression intra-class variance. Similarly negative pairs were also selected with the same gender and race to reduce the influence of other attributes. The CPLFW dataset on the contrary focuses on adding positive subjects with pose variations. They also add 3,000 positive pairs with varied poses and construct the same number of negative pairs keeping the same constraints as the CALFW dataset.

\subsection{Hyper-parameters for training and finetuning FR}
\label{sec:hyperparameters}

We utilize the following hyperparameters for training the respective face recognition model with a ResNet-50 backbone for all the datasets for consistency. We have used the same parameters for finetuning as well. We had experimented with different learning rates for the synthetic datasets but had found these parameters to be the best performing.  

\subsubsection{AdaFace}

\begin{itemize}
    \item Batch Size: 512
    \item Epochs: 26
    \item Learning rate milestones: 12, 20, 24
    \item Learning rate: 0.1 
    \item m: 0.4
    \item h: 0.333
    \item Low-resolution augmentation probability: 0.2
    \item Crop augmentation probability: 0.2
    \item Photometric augmentation probability: 0.2 
\end{itemize}

\subsubsection{ArcFace}

% Similarly, we have utilized the following hyper-parameters for training and finetuning the ArcFace model. 

\begin{itemize}
    \item Embedding size: 512
    \item Momentum: 0.9
    \item Weight Decay: 5e-4
    \item Batch Size: 128
    \item Learning rate: 0.02
    \item Epochs: 20
    \item Margin list: (1.0, 0.5, 0.0)
\end{itemize}

\subsubsection{ElasticFace}

\begin{itemize}
    \item Epoch: 40
    \item Batch size: 128
    \item Learning rate: 0.1
    \item s: 64.0
    \item m: 0.5
    \item std: 0.0175
    \item Momentum: 0.9
    \item Warmup: -1
    \item Weight decay: 5e-4
    \item Embedding size: 512
\end{itemize}

% \section{Experimentation with Other Approaches}
% \label{sec:other_approaches}

% \section{Results on the Bupt-GloablFace Dataset}

% \section{Examples of Ge}

% \begin{figure*}
    
% \end{figure*}
\begin{figure*}
    \centering
    \includegraphics[width=\linewidth,trim={12cm 12cm 12cm 12cm},clip]{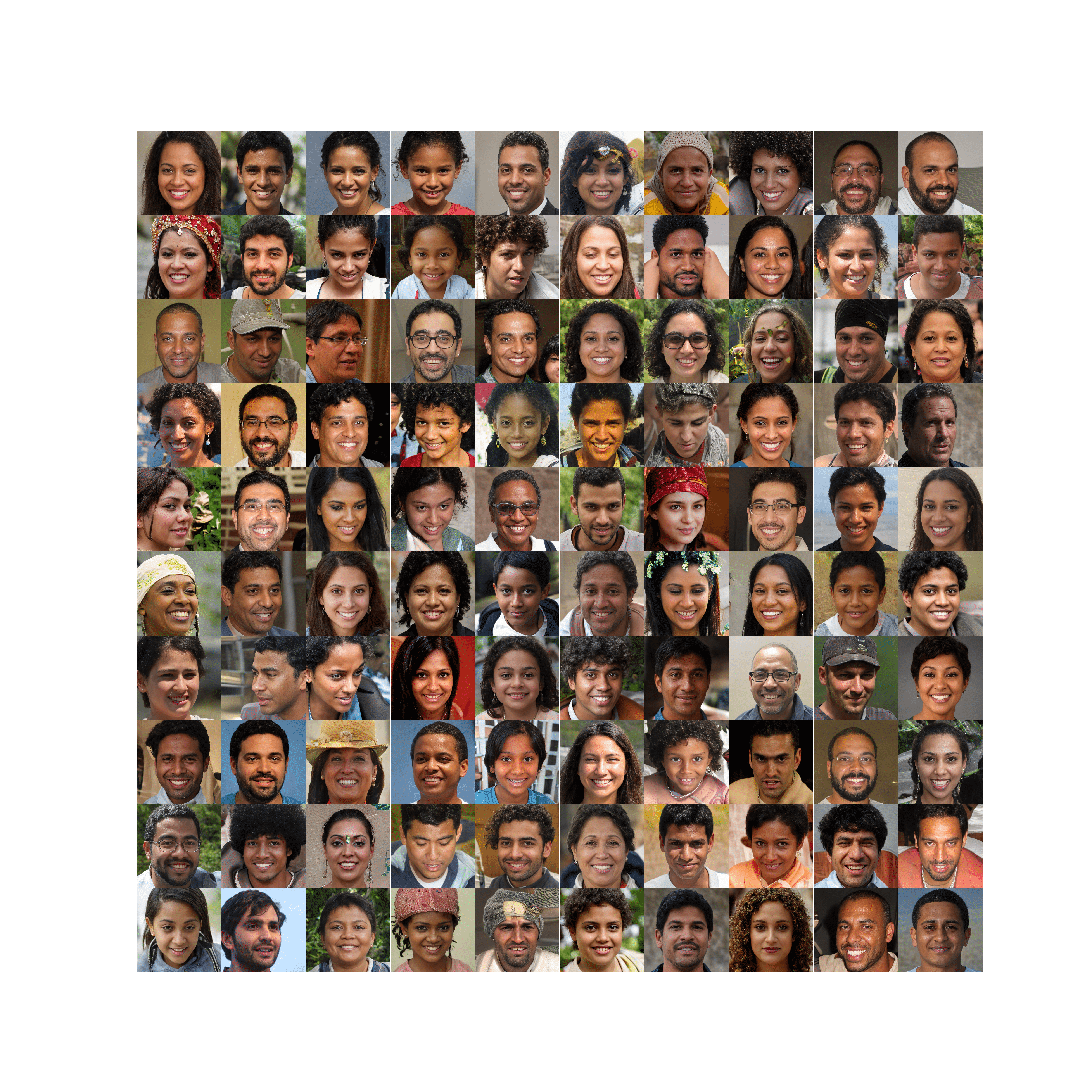}
    \caption{Examples of 100 different identities corresponding to the "Indian" racial group generated using the proposed approach. These have been randomly selected from the dataset showing different poses and expression.}
    \label{fig:enter-label}
\end{figure*}

\begin{figure*}
    \centering
    \includegraphics[width=\linewidth,trim={12cm 12cm 12cm 12cm},clip]{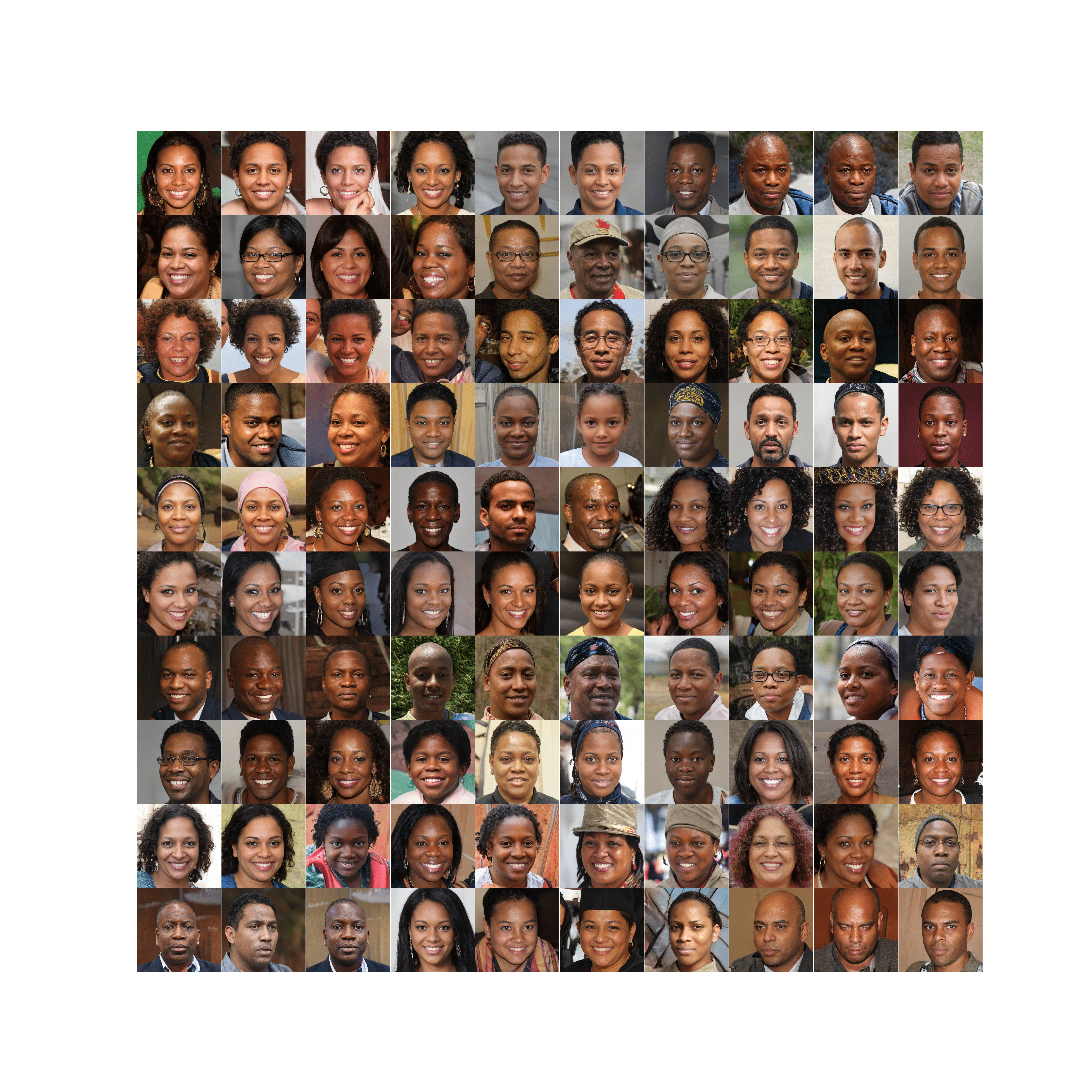}
    \caption{Examples of 100 different identities corresponding to the "African" racial group generated using the proposed approach. These have been randomly selected from the dataset showing different poses and expression.}
    \label{fig:enter-label}
\end{figure*}

\begin{figure*}
    \centering
    \includegraphics[width=\linewidth,trim={12cm 12cm 12cm 12cm},clip]{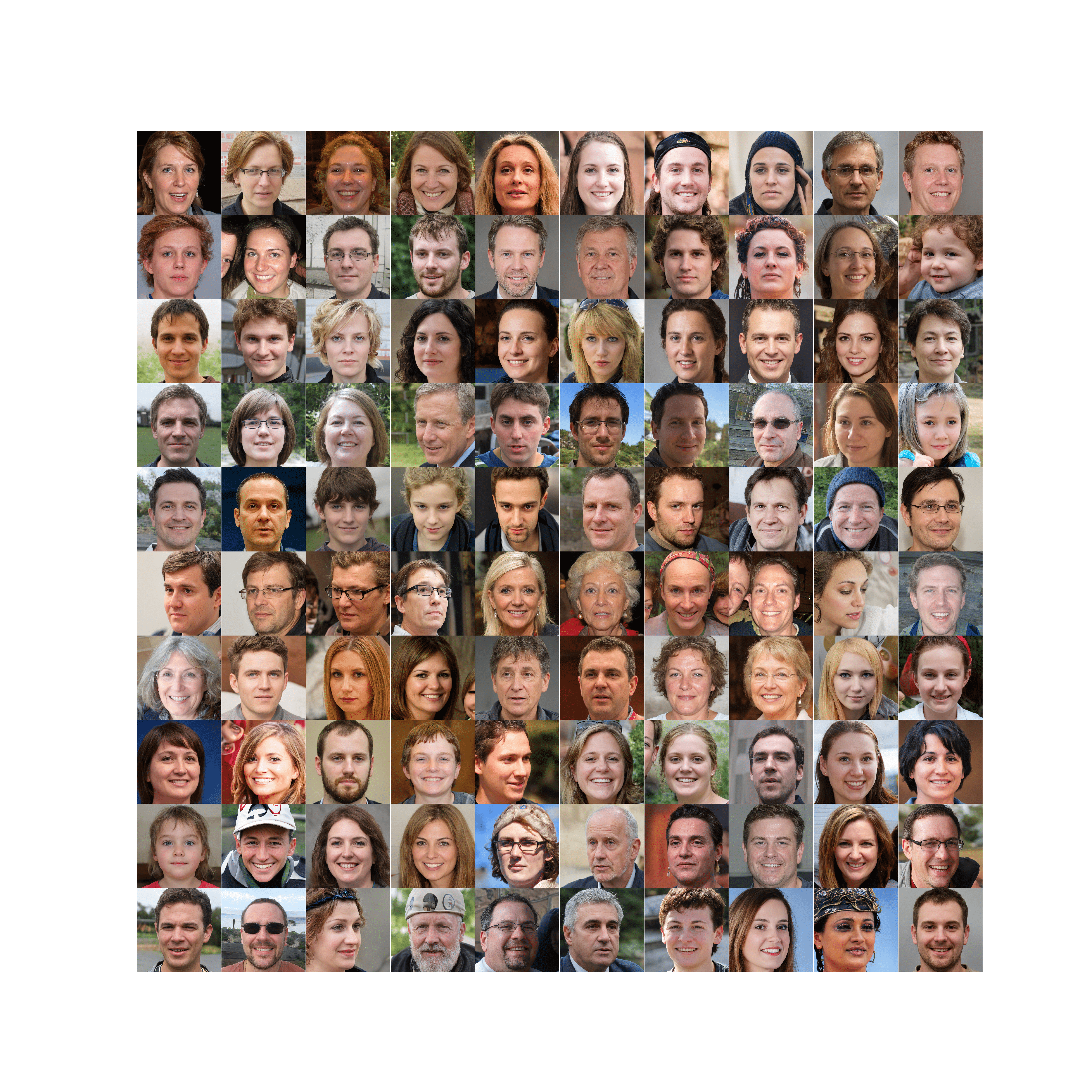}
    \caption{Examples of 100 different identities corresponding to the "Caucasian" racial group generated using the proposed approach. These have been randomly selected from the dataset showing different poses and expression.}
    \label{fig:enter-label}
\end{figure*}

\begin{figure*}
    \centering
    \includegraphics[width=\linewidth,trim={12cm 12cm 12cm 12cm},clip]{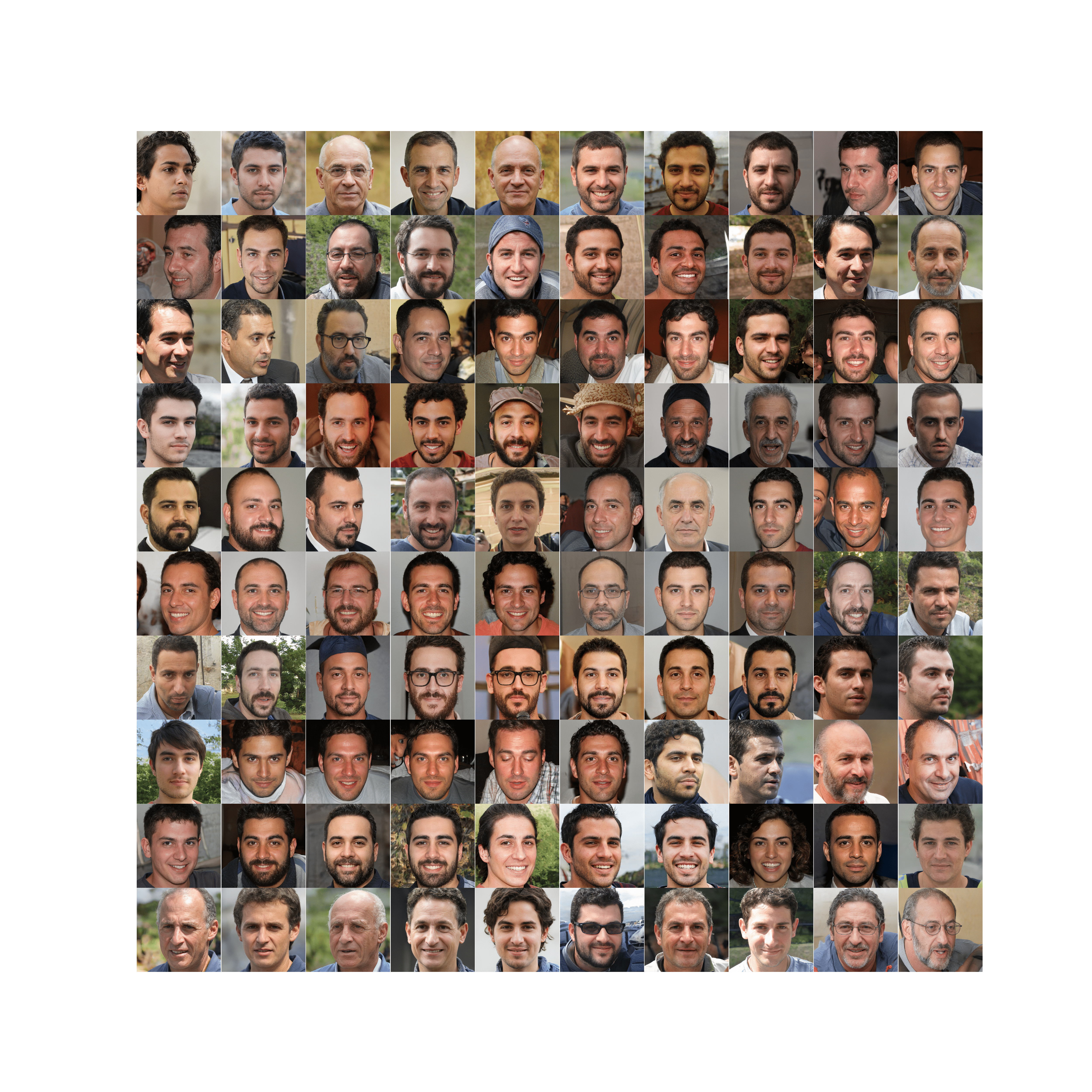}
    \caption{Examples of 100 different identities corresponding to the "Middle Eastern" racial group generated using the proposed approach. These have been randomly selected from the dataset showing different poses and expression. What is interesting is that most samples are male. There are two possible reasons - biased nature of the ethnicity classifier and the absence of female middle eastern humans in the GAN latent space. }
    \label{fig:enter-label}
\end{figure*}

\begin{figure*}
    \centering
    \includegraphics[width=\linewidth,trim={12cm 12cm 12cm 12cm},clip]{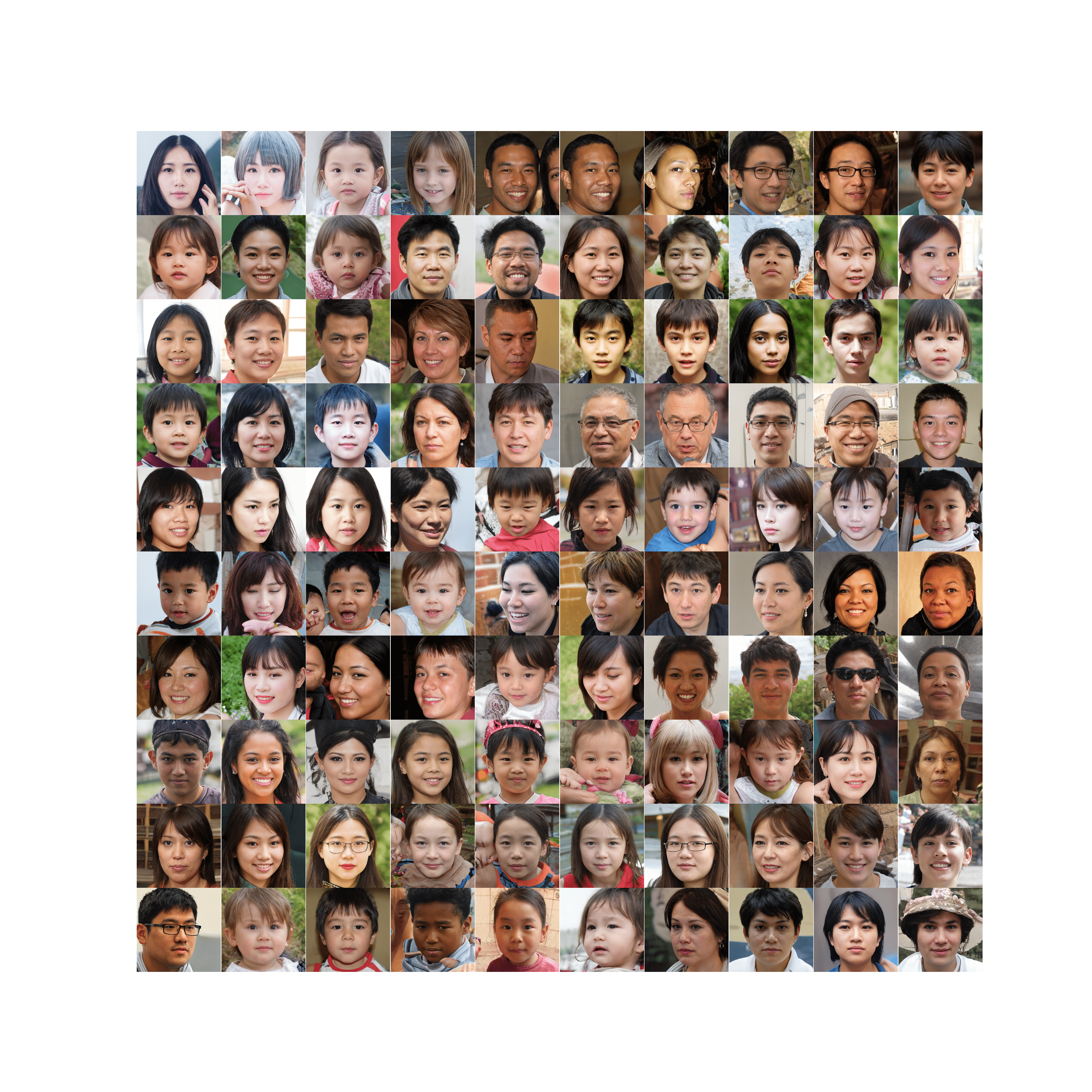}
    \caption{Examples of 100 different identities corresponding to the "Asian" racial group generated using the proposed approach. These have been randomly selected from the dataset showing different poses and expression.}
    \label{fig:enter-label}
\end{figure*}

\begin{figure*}
    \centering
    \includegraphics[width=\linewidth,trim={12cm 12cm 12cm 12cm},clip]{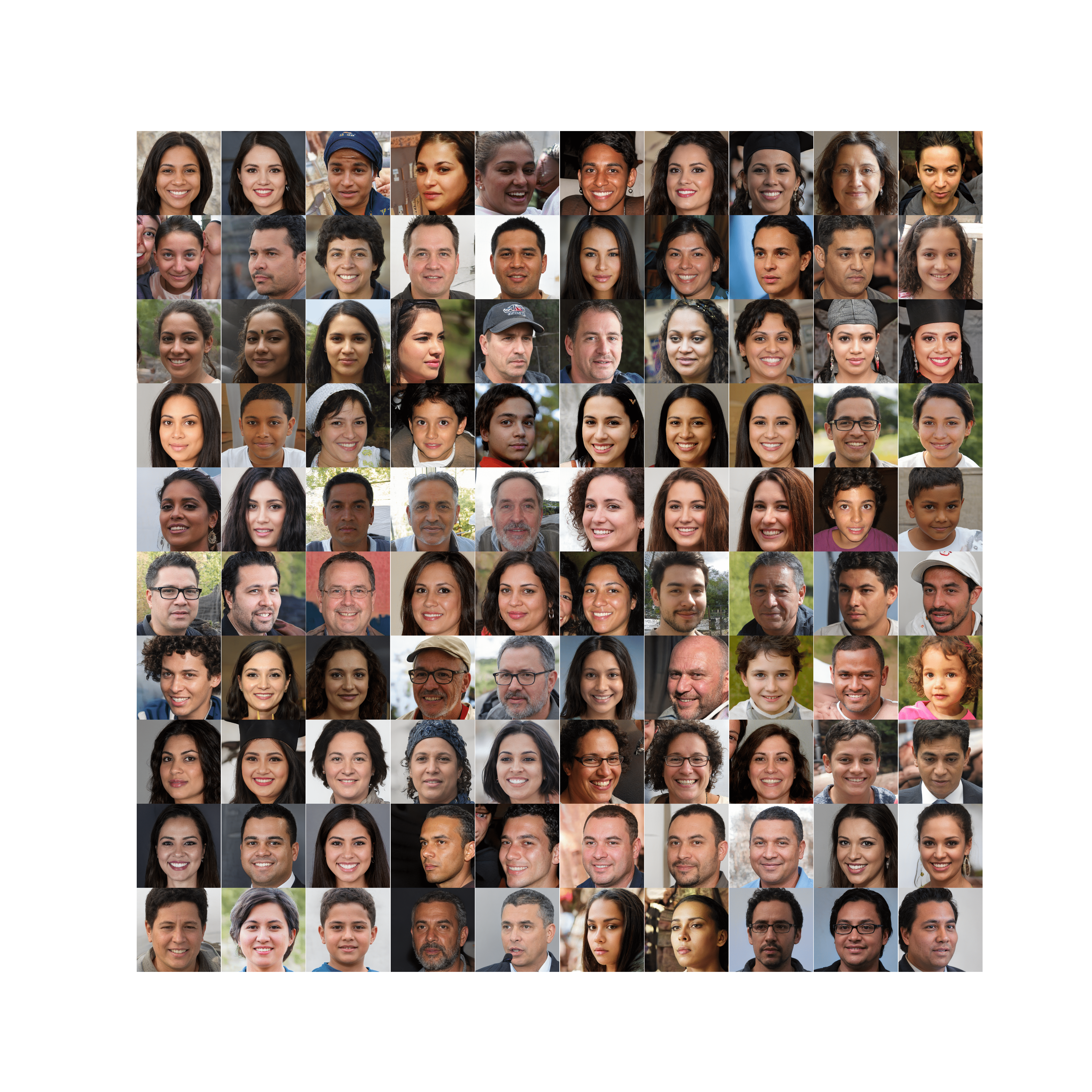}
    \caption{Examples of 100 different identities corresponding to the "Latino Hispanic" racial group generated using the proposed approach. These have been randomly selected from the dataset showing different poses and expression.}
    \label{fig:enter-label}
\end{figure*}

\end{document}